\pdfoutput=1

\documentclass[11pt]{article}

\usepackage[]{ACL2023}

\usepackage{times}
\usepackage{latexsym}

\usepackage[T1]{fontenc}

\usepackage[utf8]{inputenc}

\usepackage{microtype}

\usepackage{inconsolata}

\usepackage{tabularx}
\usepackage{graphicx}
\usepackage{multirow}
\usepackage{longtable}
\usepackage{geometry}

\newcommand{\sTag}{\texttt{<s>}}
\newcommand{\sysTag}{\texttt{\textlangle\textlangle SYS\textrangle\textrangle}}
\newcommand{\instTag}{\texttt{[INST]}}
\newcommand{\closeInstTag}{\texttt{[/INST]}}
\newcommand{\sysEndTag}{\texttt{\textlangle\textlangle /SYS\textrangle\textrangle}}

\title{Quantifying the Capabilities of LLMs across Scale and Precision}

\author{Sher Badshah \\
        Faculty of Compute Science \\
        Dalhousie University \\
        \texttt{sh545346@dal.ca} \\
        \And
        Hassan Sajjad \\
        Faculty of Computer Science \\
        Dalhousie University \\
        \texttt{hsajjad@dal.ca}}

\begin{document}
{\makeatletter\acl@finalcopytrue
  \maketitle
}
\begin{abstract}
Scale is often attributed as one of the factors that cause an increase in the performance of LLMs, resulting in models with billion and trillion parameters. One of the limitations of such large models is the high computational requirements that limit their usage, deployment, and debugging in resource-constrained scenarios. Two commonly used alternatives to bypass these limitations are to use the smaller versions of LLMs (e.g. Llama 7B instead of Llama 70B) and lower the memory requirements by using quantization. While these approaches effectively address the limitation of resources, their impact on model performance needs thorough examination. In this study, we perform a comprehensive evaluation to investigate the effect of model scale and quantization on the performance. We experiment with two major families of open-source instruct models ranging from 7 billion to 70 billion parameters.  Our extensive zero-shot experiments across various tasks including natural language understanding, reasoning, misinformation detection, and hallucination reveal that larger models generally outperform their smaller counterparts, suggesting that scale remains an important factor in enhancing performance.  We found that larger models show exceptional resilience to precision reduction and can maintain high accuracy even at 4-bit quantization for numerous tasks and they serve as a better solution than using smaller models at high precision under similar memory requirements.
\end{abstract}

\section{Introduction}
The availability of extensive data and substantial computational resources enable the pretraining of Large Language Models (LLMs) at an unprecedented scale. The increase in scale (e.g., the amount of compute budget for training, model parameters, etc.), according to a wider belief, can lead to unpredictable improvements in the performance and sampling efficiency on a broad spectrum of downstream tasks \citep{wei2022emergent, kaplan2020scaling, radford2019language, devlin2018bert}. As these models are scaled up, they begin to show emerging abilities \citep{wei2022emergent, wei2022chain}. With such an increase in size, LLMs can generate human-like text and excel in complex tasks that require creativity \citep{min2021recent, kasneci2023chatgpt, yang2023harnessing}. The alignment with methods like Reinforcement Learning from Human Feedback (RLHF) further enables a higher level of customization with human needs, thereby raising the potential applications and utility of such LLMs \citep{ouyang2022training, korbak2023pretraining}. As these models continue to improve with scale, it has now become a standard practice to train models with billions or even trillions of parameters \citep{kopf2023openassistant, balagansky2023democratized, yang2023fingpt}.

Contrary to the previous view that model performance enhances with scale which is also referred to as the scaling laws, a few studies argue that improvements do not linearly correlate with an increase in the number of parameters for certain tasks \citep{ganguli2022predictability, wei2022emergent, lin2021truthfulqa}. It also remains uncertain to what extent scaling laws apply across various downstream tasks. Moreover, achieving performance with scale carries a significant computational cost and carbon footprint. For instance, it is estimated that training GPT-3 with 175 billion parameters requires nearly 1300 megawatt-hours of electricity \citep{patterson2021carbon} and would take almost 288 years with a single NVIDIA V100 GPU \citep{narayanan2021efficient}. While it is feasible for organizations with substantial resources to train and deploy models on such an enormous scale, other entities (e.g., academic labs, general users, etc.) may experience challenges when utilizing LLMs in resource-constrained settings. For example, GPT-3 requires five NVIDIA A100 80GB GPUs to perform inference in half-precision \citep{xiao2023smoothquant}. Additionally, it can be challenging to use LLMs where high computational and communication overhead result in significant inference latency that negatively impacts user experience. In response to these challenges, techniques such as quantization have been introduced to reduce computational requirements without significantly compromising performance.

Quantization primarily involves converting the weights and activations of a neural network from their default 32-bit or 16-bit floating point formats to more compact representations such as 8-bit and 4-bit integers. Post-Training Quantization (PTQ) \citep{sung2015resiliency} is a suitable solution for addressing challenges such as extensive compute requirements and inference latency that occur during inference in a resource-constrained environment. PTQ modifies the model's weights and activations to lower precision formats without the need for retraining. While we can expect the latency and memory requirements of the model to decrease proportionally, this efficiency often comes at the cost of reduced accuracy for the end task \citep{dettmers2023case, frantar2022gptq, park2022lut}. Previous studies have suggested that 4-bit precision offers optimal scaling benefits \citep{kim2024memory, dettmers2023case}, yet it remains unclear how improvements in efficiency affect performance across various downstream tasks. This uncertainty underscores the need for a comprehensive evaluation to understand the trade-offs between performance and efficiency. Moreover, as emerging abilities appear with scale \citep{wei2022emergent, wei2022chain}, it is important to explore how quantization influences these capabilities.  

The aim of this work is to investigate the effect of scale and quantization on the performance of LLMs. We studied two major families of open-source instruct models, Llama 2~\citep{touvron2023llama} and Mistral~\citep{jiang2023mistral}, with 7 billion to 70 billion parameters. In particular, we utilized each model at different precision levels, ranging from 4-bit to 32-bit. To examine how both model families perform at these varied precision levels, we conducted comprehensive zero-shot experiments across a wide variety of tasks. We found that the \textbf{model scale tends to improve performance in most tasks}. Specifically, larger models often outperform their smaller counterparts within the same model family at similar precision. However, there are some exceptions to the benefits of scale in the reasoning tasks. For example, models perform moderately well in basic spatial reasoning but they struggle when the complexity increases. Similarly, some tasks see a decrease in performance from larger to smaller parameters. For instance, in SpartQA (hard), Mixtral 8×7B achieved a slightly lower accuracy compared to its smaller variant Mistral 7B. Furthermore, we observed that social context depends less on the scale as Mistral 7B outperformed all other models in the experiment.

Our findings on the impact of quantization revealed that \textbf{larger models are more tolerant to precision reduction} compared to their smaller counterparts. We discovered that even at 4-bit quantization, which significantly reduces memory requirements (see Table \ref{table-llama-2-chat-models-memory}), the larger models maintained high accuracy across numerous tasks. Based on our findings, we recommend that within a fixed memory budget, deploying a larger model with 4-bit quantization often yields greater benefits than utilizing a smaller model at higher precision. For instance, while a 70B model at 4-bit quantization uses only 42 gigabytes of memory—comparable to much smaller models at higher precision—it consistently delivers superior performance across a variety of tasks. This strategy effectively maximizes computational efficiency by optimizing the trade-off between memory use and model performance. 

\begin{table}
\centering
\footnotesize
\begin{tabular}{lccccc}
\hline
\textbf{Model} & \textbf{Params} & \textbf{32-bit} & \textbf{FP16} & \textbf{8-bit} & \textbf{4-bit} \\
\hline
& 7B & 56 & 28 & 14 & 7 \\
Llama 2-Chat & 13B & 104 & 52 & 26 & 13 \\
& 70B & 336 & 168 & 84 & 42 \\
\hline
\end{tabular}
\caption{Estimated GPU memory requirements (in Gigabyte) for Llama 2-Chat models at inference using various precision levels and parameter sizes \citep{kaplan2020scaling, hoffmann2022training}.}
\label{table-llama-2-chat-models-memory}
\end{table}

\section{Methodology}
This section describes the key configurations of our evaluation process: tasks, prompts, models, and quantization.

\subsection{Tasks}\label{tasks}
We utilized various datasets to evaluate the effect of scale and quantization. Table \ref{table_tasks} summarizes the tasks and datasets for evaluation, consisting of a variety of tasks including Natural Language Understanding (NLU) tasks (i.e., summarization, machine translation, and sentiment analysis), reasoning, hallucination, and misinformation detection tasks. Due to the limited computing resources, we adapted a sampling approach of \citep{bang2023multitask} and considered 30 to 660 test samples for each task. To evaluate the model-generated responses, we performed automated evaluation on standard NLU tasks. Subsequently, we assessed reasoning, hallucination, and misinformation detection tasks through human evaluation. Appendix \ref{app:tasks} provides a detailed explanation of each task along with the number of selected samples and the evaluation strategy.

\subsection{Prompt Making}
Our evaluation protocol assesses the model capabilities on all tasks under a zero-shot setting, without any examples or chain of thought prompting \citep{wei2022chain}. This approach allows for a more direct assessment of the intrinsic abilities of the models. We incorporate various prompting strategies (see Table \ref{table-prompting-strategies}) to elucidate the extent to which the difference in input may influence the performance and behavior of the models under study.  Our preliminary experimentation revealed that role-playing \citep{kong2023better} is particularly effective when combined with other prompting techniques given in Table \ref{table-prompting-strategies}. Therefore, we used a combination of role-playing, templated \citep{touvron2023llama, jiang2024mixtral}, and direct-to-detail prompting (see Appendix \ref{app:prompting} for examples).

\subsection{Models}
We evaluate two major open-source LLM families: Llama 2-Chat~\citep{touvron2023llama} and Mistral Instruct models~\citep{jiang2023mistral}. Both are decoder-only models. Llama 2-Chat includes variants with 7 Billion (7B), 13 Billion (13B), and 70 Billion (70B) parameters. It incorporates supervised fine-tuning and RLHF methods such as proximal policy optimization and rejection sampling to refine and improve dialogue use cases and responsible AI \citep{touvron2023llama}. In our evaluation work, we considered 7B and 70B variants to understand how varying model sizes or parameter scaling affect performance. On the other hand, Mistral Instruct models are fine-tuned to follow instructions. Mistral 7B Instruct is a fine-tuned version of Mistral 7B that employs grouped query and sliding window attentions for improved efficiency and performance \citep{jiang2023mistral}. Similarly, Mixtral 8\(\times \)7B Instruct is a chat model to follow instructions using supervised fine-tuning and direct preference optimization \citep{jiang2024mixtral}. We experimented with two specific versions of the Mistral Instruct models: Mistral-7B-Instruct-v0.2 and Mixtral-8x7B-Instruct-v0.1. For consistency, we will refer to the models as Mistral 7B and Mixtral 8\(\times \)7B throughout the remainder of the paper.

\subsection{Quantization}
We used LLM.int8() \citep{dettmers2022gpt3} for 8-bit quantization. LLM.int8() is a vector-wise quantization technique that employs mixed-precision quantization to retain outlier submatrices in FP16 and standard submatrices in INT8. This mixed-precision approach allows for separate computations of FP16 outlier and INT8 non-outlier submatrices which are then combined to maintain computational efficiency and precision. Consequently, LLM.int8() effectively balances between reducing model size and preserving important data features. For 4-bit quantization, we employed QLoRA \citep{dettmers2024qlora}, as it utilizes a high-precision 4-bit NormalFloat (NF4) quantization method alongside Low-rank Adapters. This technique allows for maintaining high computational precision with compact 4-bit storage. QLoRA effectively balances precision and efficiency in a resource-optimized manner.

\subsection{Experimental Settings}
We utilized \texttt{bitsandbytes} library \citep{dettmers2024qlora, dettmers2023case, dettmers2022llm} to quantize each model to 4 and 8-bit.  For half-precision (FP16), we leveraged PyTorch's capabilities to work with lower-precision arithmetic directly. This is accomplished through the use of the \texttt{torch.float16} data type \citep{paszke2019pytorch} that allows the opportunity to experiment with half-precision floating-point numbers. For comparison, we established two baselines: models operating under full precision using 32-bit floating-point (FP32) and using half-precision (FP16). We set the temperature value to 0.6, a repetition penalty of 1.2, a top-k value of 50, and a top-p value of 0.9. The batch sizes are tailored to each model variant: a batch size of 8 for the 7 billion parameter models and a batch size of 2 for other model variants.

\begin{table*}
\centering
\begin{tabularx}{\textwidth}{llX}
\hline
\textbf{Tasks} & \textbf{Datasets} & \textbf{Reference} \\
\hline
Deductive & EntailmentBank, bAbI (Task 15) & \citep{dalvi2021explaining, weston2015towards} \\
Inductive & CLUTRR, bAbI (Task 16) & \citep{sinha2019clutrr, weston2015towards} \\
Abductive & $\alpha$NLI & \citep{bhagavatula2019abductive} \\
Temporal & TimeDial & \citep{qin2021timedial} \\
Spatial & SpartQA, StepGame & \citep{mirzaee2021spartqa, shi2022stepgame} \\
Mathematical & MATH & \citep{saxton2019analysing} \\
Commonsense & CommonsenseQA, PiQA, Pep-3k & \citep{talmor2018commonsenseqa, bisk2020piqa, wang2018modeling} \\
Causal & e-CARE & \citep{du2022care} \\
Multi-hop & HotpotQA & \citep{yang2018hotpotqa} \\
Analogical & Letter String Analogies & \citep{webb2023emergent} \\
Hallucination & TruthfulQA & \citep{lin2021truthfulqa} \\
Misinformation detection & COVID fact-checking & \citep{lee2021towards} \\
Summarization & CNN/Daily Mail, SAMSum & \citep{hermann2015teaching, gliwa2019samsum} \\
Machine Translation & FLoRes-200 & \citep{costa2022no} \\
Sentiment Analysis & NusaX & \citep{winata2022nusax} \\
\hline
\end{tabularx}
\caption{Tasks and corresponding datasets for evaluation.}
\label{table_tasks}
\end{table*}

\begin{table*}[h]
\centering
\begin{tabular}{p{0.3\linewidth}p{0.6\linewidth}}
\hline
\textbf{Strategy Type} & \textbf{Description} \\ 
\hline
Role-playing & Models assume predefined roles, such as a sentiment analysis assistant, to provide context-specific responses \citep{kong2023better}. \\
Templated Prompting & Structured instructions are embedded within a template to ensure consistent and safe interactions across tasks. This includes directives to be helpful, respectful, and honest, as well as avoiding harmful or biased content \citep{touvron2023llama, jiang2024mixtral}. \\
Direct to Detail Prompting & Prompts range from minimal guidance, providing direct instructions, to detailed guidance, specifying constraints such as word limits and content restrictions to shape the response. \\
\hline
\end{tabular}
\caption{Overview of prompting strategies employed}
\label{table-prompting-strategies}
\end{table*}

\section{Results and analysis}
We observed comparable performance between the FP16 and FP32 models. Consequently, in subsequent analyses, we will designate the FP16 models as the baseline for comparison with the 8-bit and 4-bit quantized models (see Appendix \ref{app:additional_results} for details).

\subsection{Reasoning}
\begin{figure*}[ht]
\centering
\includegraphics[width=0.7\textwidth]{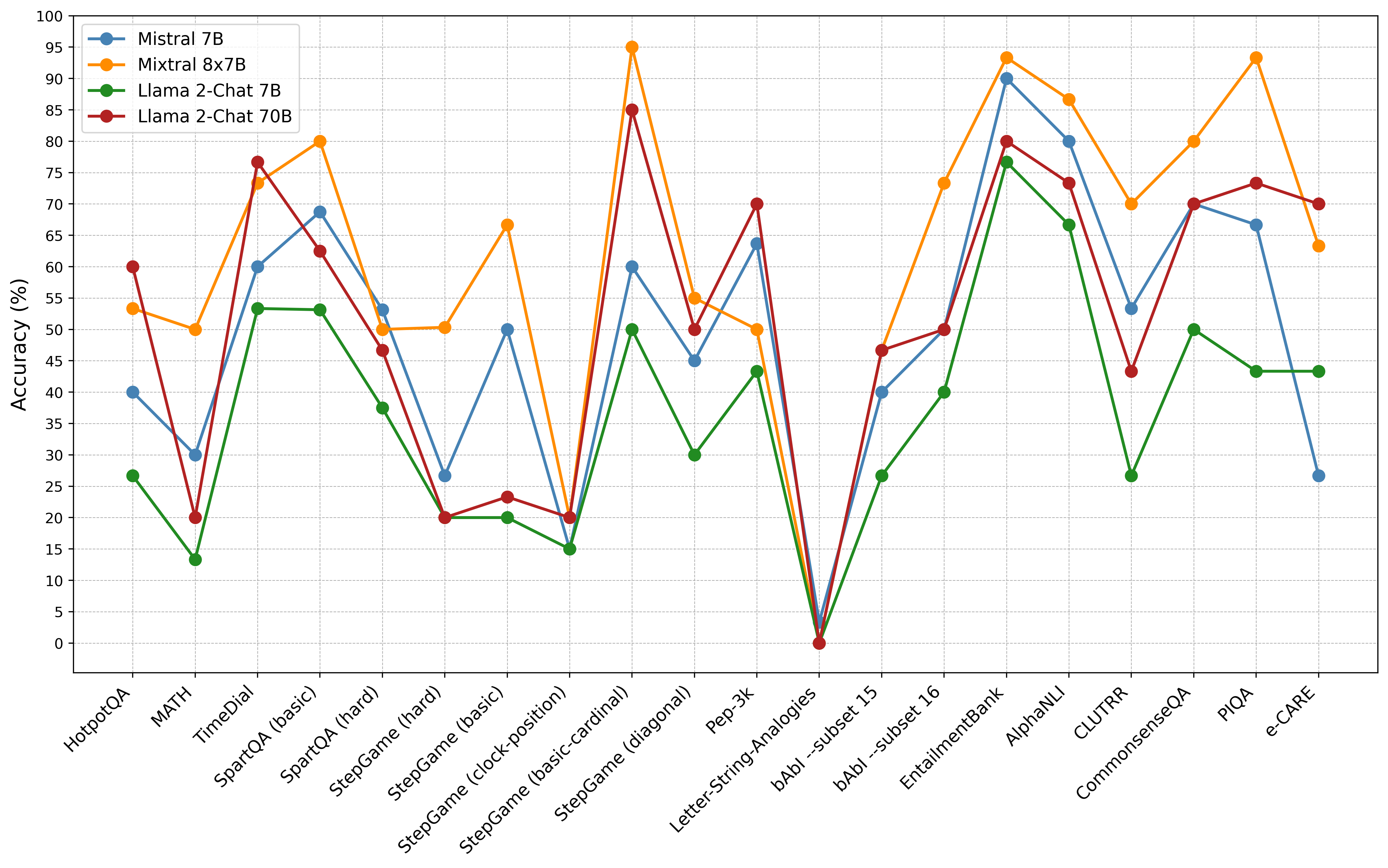}
\caption{Performance of Llama 2-Chat and Mistral models across reasoning tasks operating under FP16 precision}
\label{fig:scaling_effect_tasks}
\end{figure*}

Our evaluation of both model families across various reasoning tasks reveals significant insights into the impact of model scaling and quantization on reasoning capabilities. From Figure \ref{fig:scaling_effect_tasks}, it is evident that the scale influences the performance of the models on reasoning tasks. Particularly, in both model families, \textbf{the larger model often outperforms its smaller counterpart} which aligns with the findings of previous studies that larger models tend to perform better on complex reasoning tasks \citep{wei2022emergent}. The improvement with scale is due to the capacity of larger models to capture more complex patterns and dependencies in the data \citep{kaplan2020scaling}. This is particularly evident in tasks such as StepGame (basic cardinal) \citep{shi2022stepgame} and EntailmentBank \citep{dalvi2021explaining}. However, we also noted that the scale does not consistently lead to better performance.

The difference in performance across reasoning tasks presents the complex nature of human cognitive processes and the challenge of emulating these through LLMs. In tasks such as analogical reasoning (i.e., Letter string analogies), even the largest models failed to perform. This shows a potential gap in the model's ability to handle abstract reasoning and suggests that the current scaling methods do not inherently equip models with the ability to handle the complexity of such tasks. Tasks requiring temporal and commonsense reasoning show relatively high accuracy, revealing that \textbf{larger models are particularly proficient at tasks that need integrating contextual knowledge and understanding of everyday logic}. On the other hand, spatial reasoning presents an interesting case; while the models perform moderately well on basic spatial reasoning (i.e., SpartQA~\citep{mirzaee-etal-2021-spartqa}), they struggle when the complexity increases as can be seen in StepGame (hard) \citep{shi2022stepgame}.

To summarize, while larger scales tend to enhance reasoning performance, the relationship between scale and reasoning is not straightforward. The increase in parameters appears to afford the models a more comprehensive understanding of context, logical structures, and causal relationships. However, this does not uniformly translate to better performance across all reasoning types, particularly when it comes to analogical and mathematical reasoning, suggesting a different approach to learn these tasks that are beyond scaling. 

\begin{figure}[ht]
\centering
\includegraphics[width=\linewidth]{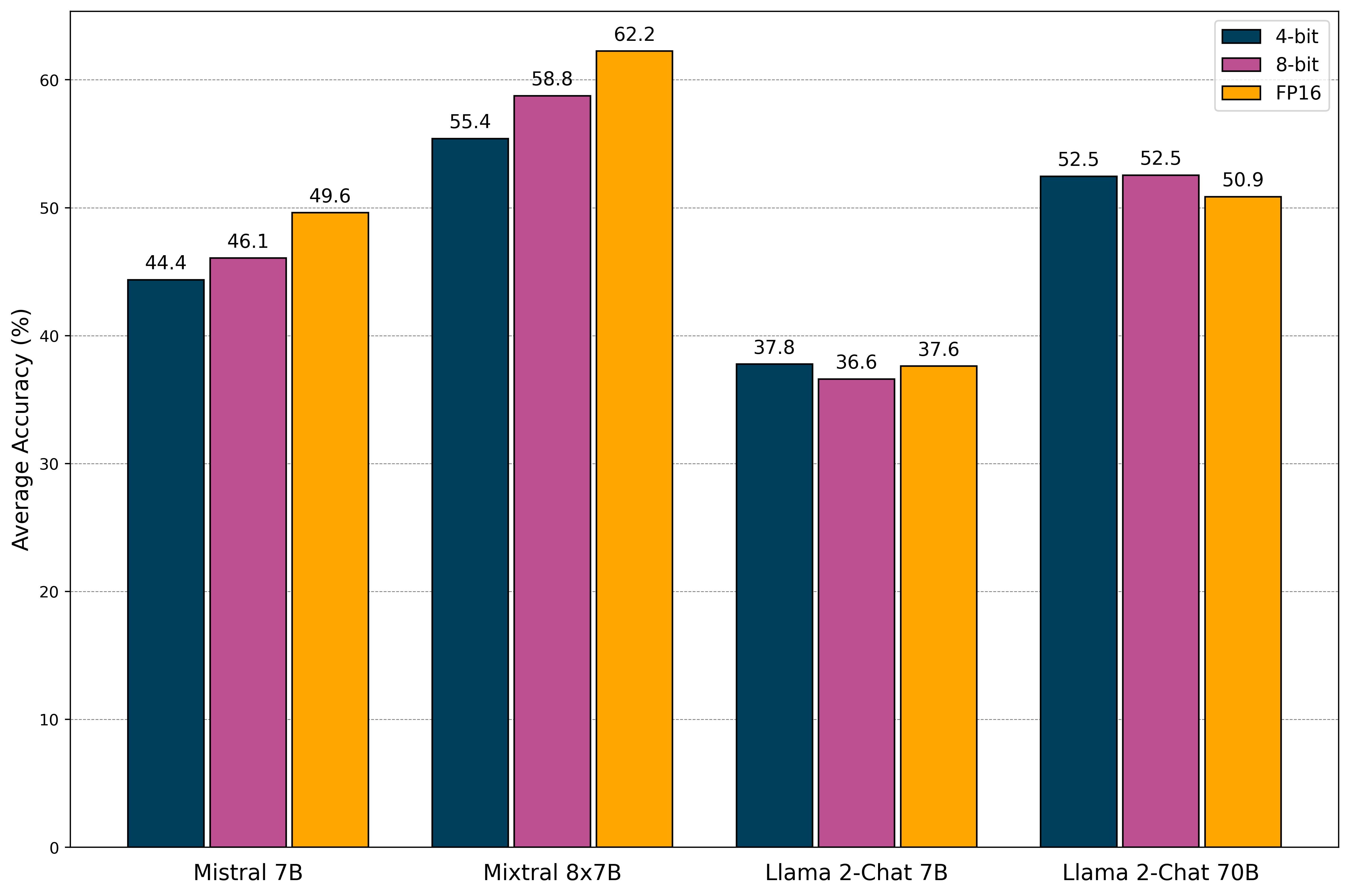}
\caption{Effect of 4 and 8-bit quantization on models reasoning compared to half-precision}
\label{fig:reasoning_quantization_avg}
\end{figure}

Figure \ref{fig:reasoning_quantization_avg} provides a clear perspective on the efficacy of both open-source model families when operated under various precision levels. Contrary to the explicit expectation that higher precision correlates to superior performance, the data suggests a more complex reality where \textbf{lower precision does not uniformly degrade performance and in some instances, seems to have an unexpectedly minimal impact on performance.}

Across all reasoning tasks, the average performance indicates that Llama 2-Chat models are less impacted by 4-bit and 8-bit quantization. In contrast, both Mistral 7B and  Mixtral 8\(\times \)7B experience a slight decrement in performance as the models are scaled down to 4 and 8 bits. This differential response highlights the variability in how different model architectures or pre-training approaches cope with the constraints imposed by quantization. The slight performance variations across both families at reduced precision levels suggest that precision downscaling can be a viable path toward computational efficiency without substantial sacrifices in reasoning capabilities.

We examined the effect of quantization on each reasoning task. In the context of mathematical reasoning, the performance appears relatively unaffected by precision, with 4-bit maintaining a similar accuracy to that of F16 across all model sizes. For tasks such as TimeDial \citep{qin2021timedial} and EntailmentBank \citep{dalvi2021explaining}, where models are expected to determine and reason over fine-grained temporal sequences and logical steps, one might predict a pronounced drop with lower precision. However, the results do not uniformly reflect this. Particularly for the 70B model, there is notable maintenance of high accuracy even at reduced precision that suggests the model's understanding of these complex concepts is encoded in a way that is resilient to precision downscaling. Interestingly, for StepGame (basic and hard) \citep{shi2022stepgame}, there is a small improvement in accuracy at 4-bit compared to F16 in the Llama 2-Chat 7B model. It is also worth noting that certain tasks such as the bAbI dataset \citep{weston2015towards} present a mixed response to changes in precision, with some model sizes showing sensitivity while others do not. This inconsistency could reflect the diverse nature of reasoning required for these tasks, where some aspects may rely more on precision than others. Appendix~\ref{app:additional_results} includes the performance of each reasoning task across 4-bit, 8-bit, FP16, and FP32.

\subsection{Hallucination and Misinformation}
\begin{figure}[ht]
\centering
\includegraphics[width=\linewidth]{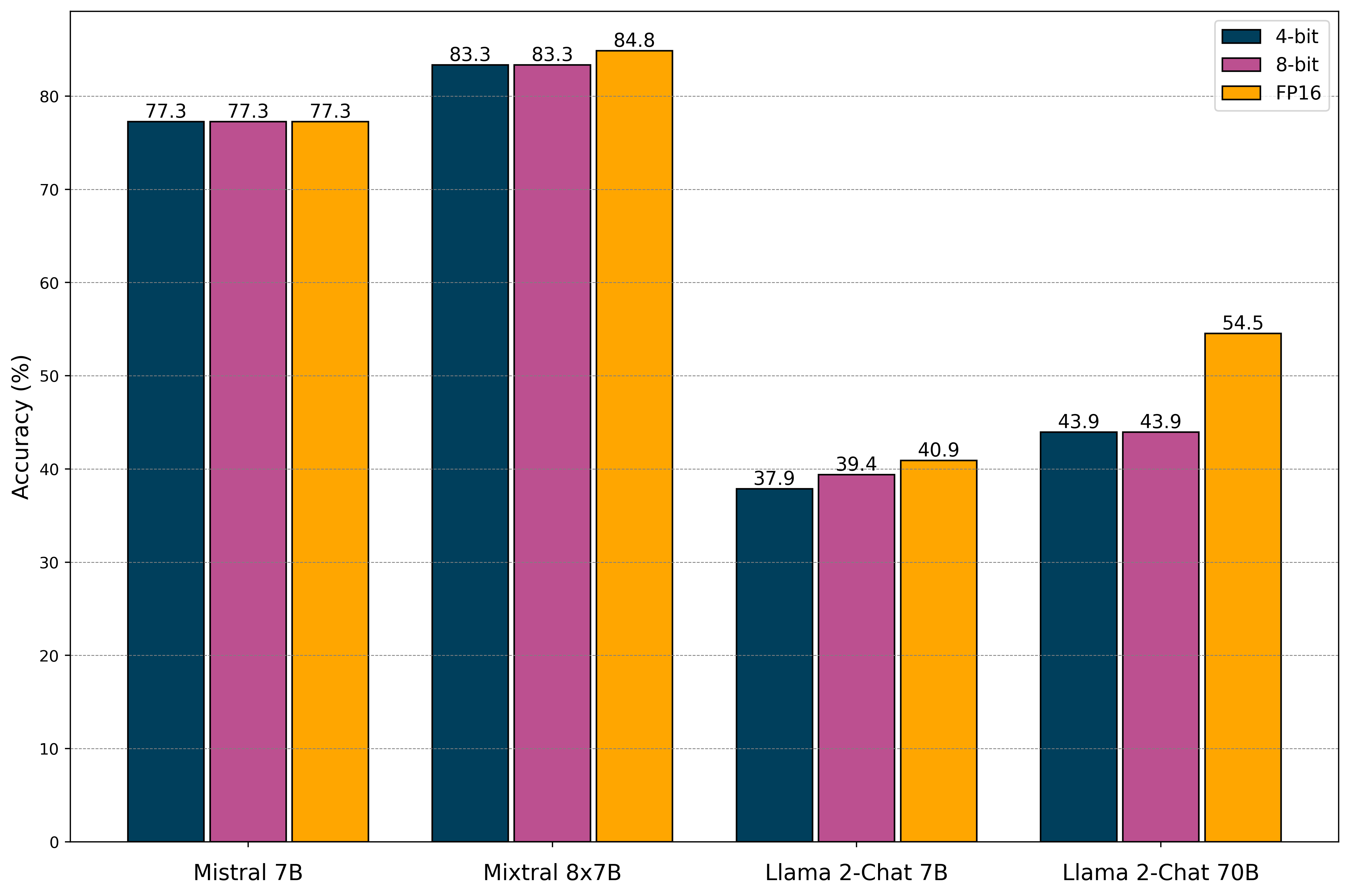}
\caption{Performance of Mistral and Llama 2-Chat models on TruthfulQA~\citep{lin2021truthfulqa} at scale and precision}
\label{fig:hallucination}
\end{figure}

Across both model families, we found that \textbf{larger models are more truthful}. As illustrated in Figure \ref{fig:hallucination}, Mixtral 8\(\times \)7B and Llama 2-Chat 70B outperformed their smaller variants. This improvement challenges the previously held belief associated with the Inverse Scaling Law (ISL) \citep{mckenzie2023inverse} that larger models are inherently less truthful \citep{lin2021truthfulqa}. During scaling, the model size significantly increases. Our findings suggest that the increase in model size does not adhere to the expectations of ISL. Rather, the performance of larger models deviates from ISL. 

Figure \ref{fig:hallucination} illustrates that larger models in both model families exhibit comparable performance in 4 and 8-bit quantization. In contrast to our findings in reasoning tasks, where the Llama model family showed tolerance towards quantization, the same model family performance on TruthfulQA \citep{lin2021truthfulqa} reveals a marked sensitivity to higher precision. As depicted in Figure~\ref{fig:hallucination}, the 70B model performance substantially increases from 43.94\% at 8-bit quantization to 54.55\% when utilized in FP16.

\begin{figure}[ht]
\centering
\includegraphics[width=\linewidth]{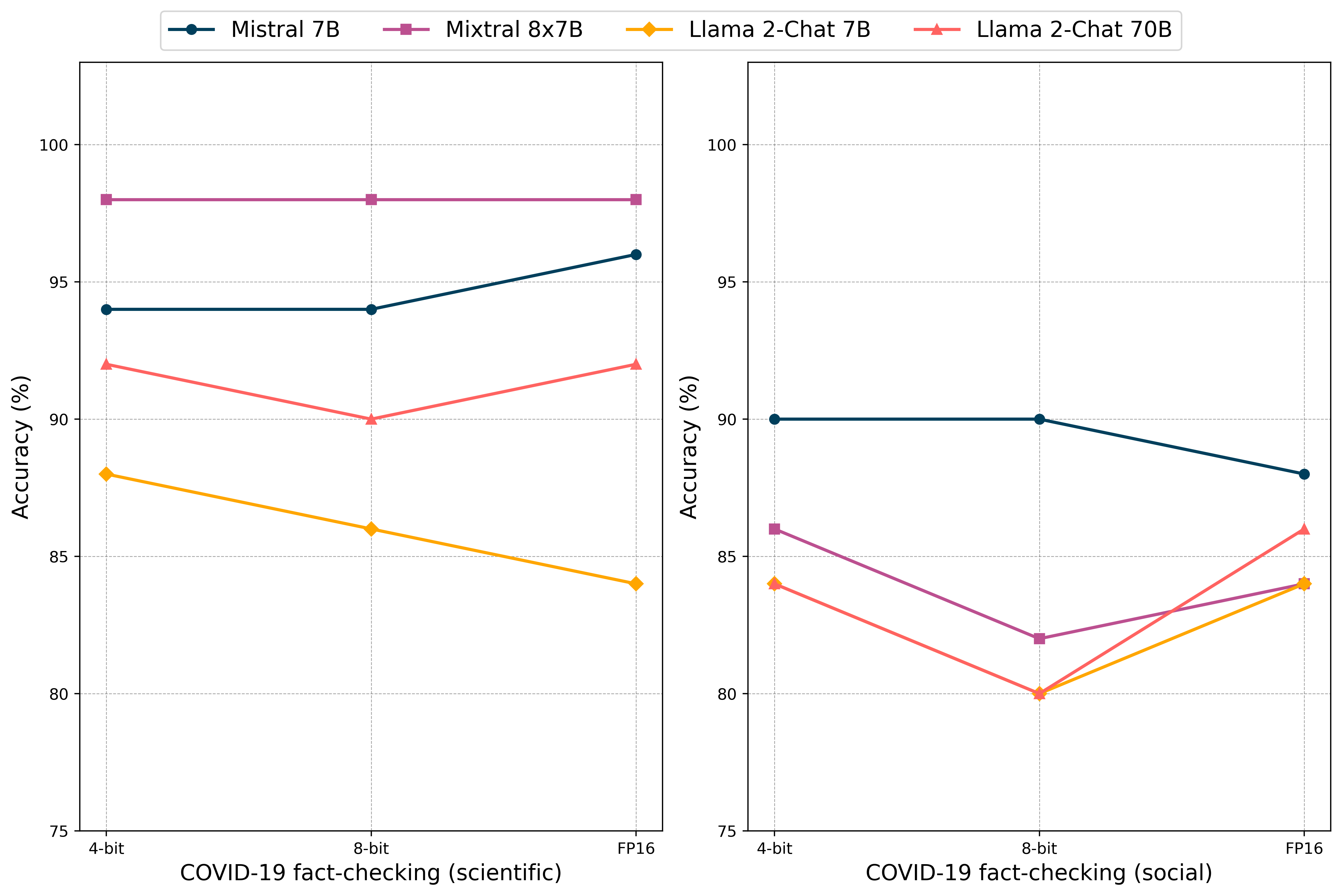}
\caption{Performance of both model families on COVID-19 fact-checking \citep{lee2021towards} across precisions}
\label{fig:misinformation}
\end{figure}

In the COVID-19 fact-checking task~\citep{lee2021towards}, \textbf{larger models within both families are better at detecting scientific misinformation}. For example, as given in Figure \ref{fig:misinformation}, the  Mixtral 8\(\times \)7B model showed outstanding performance in a scientific subset and outperformed its smaller variant. Similarly, in the Llama 2-Chat model family, the larger 70B exceeded 7B in detecting scientific fallacies. The analysis also revealed that \textbf{smaller models are more sensitive to quantization} such as Llama 2-Chat 7B consistently dropped its accuracy score from 88 at 4-bit to 84 at FP16. In across model families comparison, Mistral achieved greater accuracy compared to Llama 2-Chat in the scientific subset. However, we observed different performance patterns from both model families in the social subset. As depicted in the social plot of Figure \ref{fig:misinformation}, \textbf{smaller models are more accurate at detecting social myths}. The Mistral 7B outperformed the larger  Mixtral 8\(\times \)7B. Similarly, 7B and 70B in the Llama 2-Chat perform comparable performance in 4 and 8-bit quantization. Nevertheless, the accuracy of Llama 2-Chat 70B is slightly better in FP16. 

In summary, larger models often perform poorly on tasks requiring an understanding of social contexts and norms \citep{solaiman2019release}. Another promising aspect we observed is that \textbf{social context depends less on the scale.} This is particularly evident in the Mistral 7B model which achieved better accuracy than the  Mixtral 8\(\times \)7B and Llama 2-Chat 70B model. In summary, quantization has a variable impact on the performance of both model families in this specific setting. While the larger models typically have higher accuracy, this is inconsistent across all quantization levels, particularly in the social subset where the performance differences can diminish or even reverse.

\subsection{Natural Language Understanding}
The performance of evaluated models varies across CNN/Daily Mail \citep{hermann2015teaching} and SAMSum \citep{gliwa2019samsum} datasets. The results demonstrate that the models achieved higher ROUGE-1 scores on the SAMSum dataset. Interestingly, while the structured news articles of the CNN/Daily Mail typically require an understanding of complex narrative structures, the informal and conversational nature of SAMSum resonates more effectively with the Llama 2-Chat 70B and  Mixtral 8\(\times \)7B. This allows produce summaries closer in quality to human-generated benchmarks.

\begin{figure}[ht]
\centering
\includegraphics[width=\linewidth]{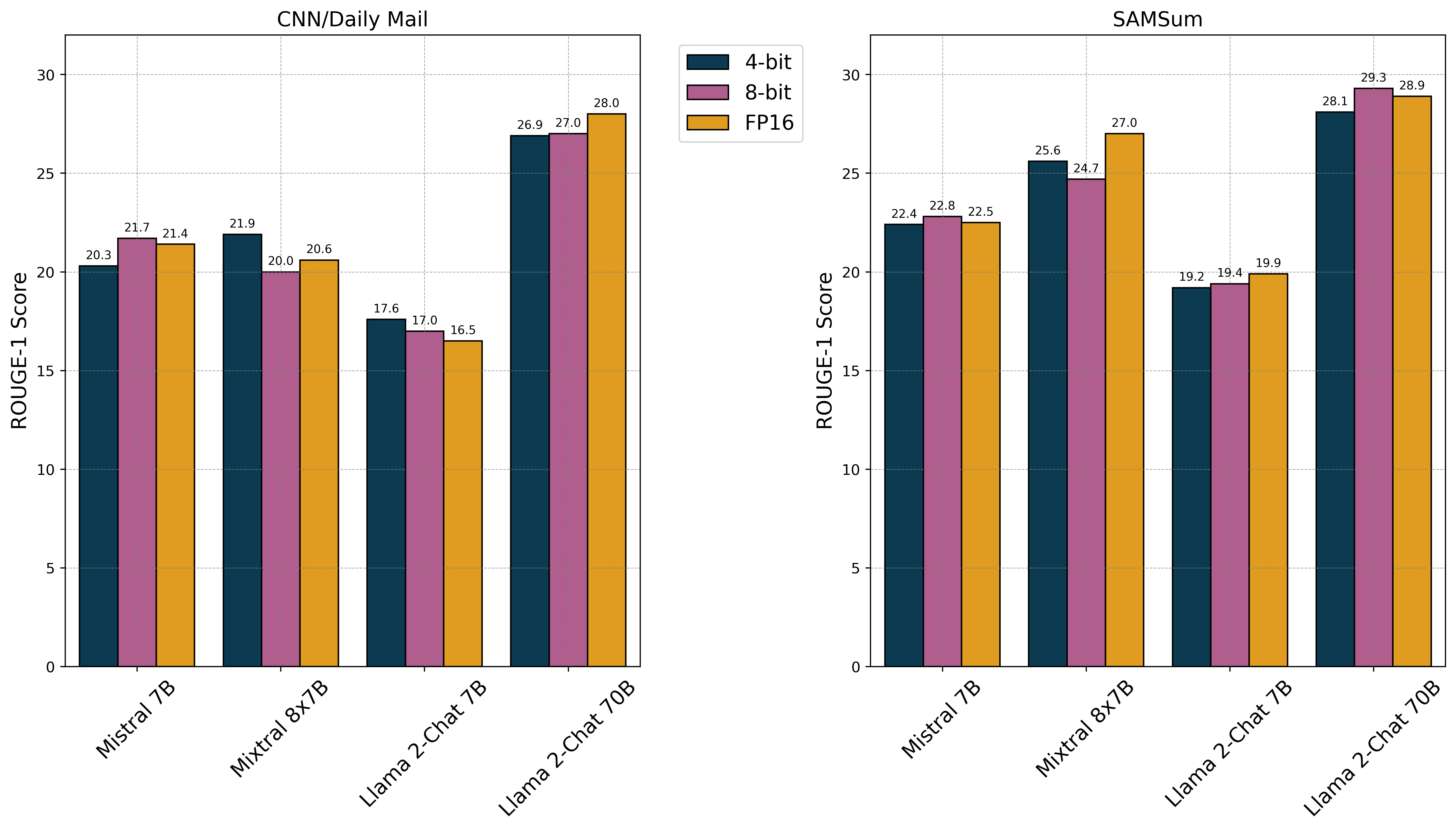}
\caption{ROUGE-1 scores of Llama 2-Chat and Mistral models on summarization tasks in different precisions}
\label{fig:summarization}
\end{figure}

In Figure \ref{fig:summarization}, it is evident that the larger Llama 2-Chat 70B consistently outperforms their smaller counterparts in achieving higher scores. Despite the variations in computational precision, the 70B model showed an impressive ability to maintain high-quality summarization performance. This observation underscores the hypothesis that \textbf{increasing the model size enhances natural language understanding}  \citep{rae2021scaling, kaplan2020scaling}. Even when operating at reduced precision levels such as 4-bit and 8-bit, the model ROUGE-1 scores remained robust. However, the performance trends across different quantization levels in both model families suggest that \textbf{the advantage of larger scale is not uniformly experienced across all computational precisions}. More specifically, while the Llama 2-Chat 70B model demonstrates notable resilience at lower precision levels, the variations in performance highlight a complex interplay between scale, architectural efficiency, and quantization. Furthermore, Mistral models introduce a fascinating perspective on the role of architectural design decisions alongside scale and quantization. The Mistral 7B and  Mixtral 8\(\times \)7B models show consistency and adaptability across precision levels. The Mistral 7B achieved almost identical performance across all precision levels. However,  Mixtral 8\(\times \)7B shows higher sensitivity to quantization in the SAMSum task.

\begin{figure}[ht]
\centering
\includegraphics[width=\linewidth]{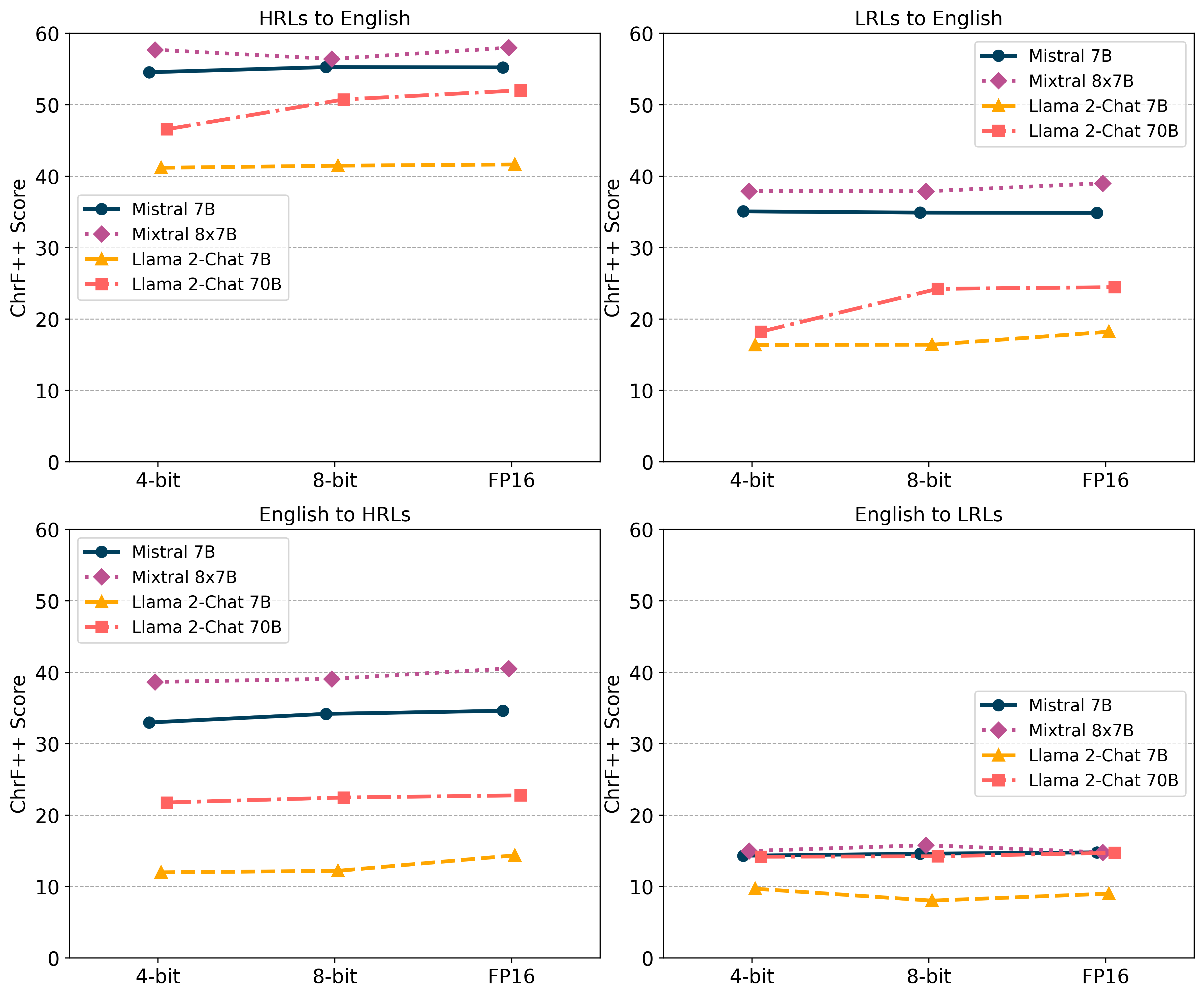}
\caption{Llama 2-Chat and Mistral machine translation performance across different precisions}
\label{fig:translation}
\end{figure}

The machine translation results in Figure \ref{fig:translation} show that models within the Mistral family surpass Llama 2-Chat in the experiment and obtained nearly matching performance across quantization and half-precision. In Llama 2-Chat, there is a slight drop in translation accuracy at lower precision levels, yet, the decrease is not as severe as anticipated. The larger models in our experiment, particularly those belonging to the \textbf{Mistral family show resilience to precision reduction}. Interestingly, this trend persists even as the precision is scaled down from FP16 to 4-bit quantization. Our experiments across language pairs reveal that the performance gains associated with \textbf{larger models are more pronounced when translating between English and Low Resource Languages (LRLs) compared to High Resource Languages (HRLs)}. 

\begin{figure*}[ht]
\centering
\includegraphics[width=1\textwidth]{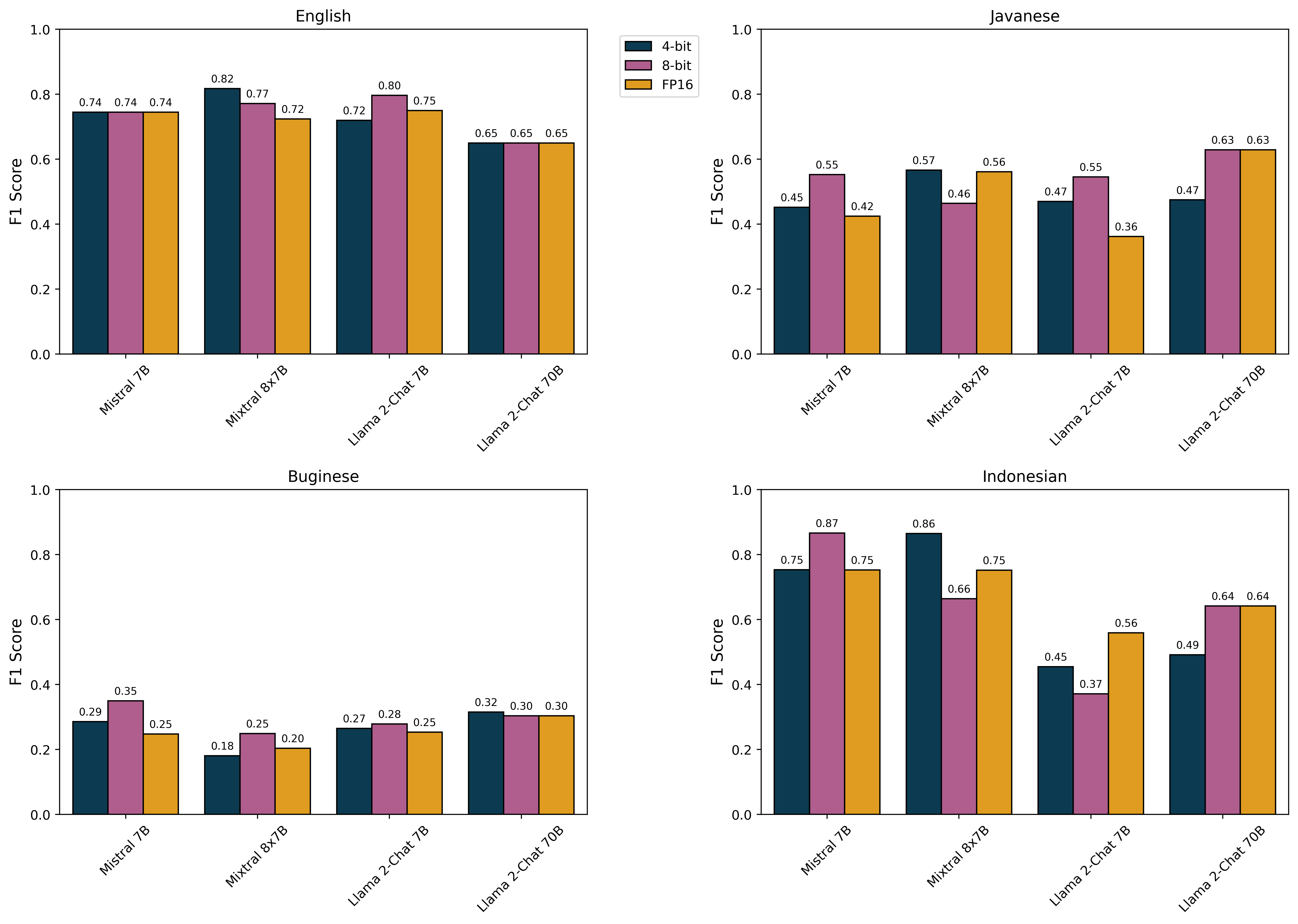}
\caption{Performance on NusaX~\citep{winata2022nusax} at different scales and precisions}
\label{fig:sa} 
\end{figure*}

In the sentiment analysis task, we observed a varied pattern. The larger Llama 2-Chat 70B performs worse than the other models in the experiment for English (see Figure \ref{fig:sa}). However, its smaller variant, Llama 2-Chat 7B, shows nearly similar performance to that of  Mixtral 8\(\times \)7B and Mistral 7B in the same language category. We found that the evaluated models specifically struggle with Buginese and show distinct results across various precision levels. Nonetheless, the difference in performance between 4-bit and 8-bit quantization and FP16 is minimal.

\section{Related Work}
Recent years have witnessed an increasing interest in the evaluation of LLMs. The key objective of the evaluation is to understand the current capabilities and limitations of LLMs and provide improvement recommendations. In the LLMs evaluation, key contributions include the introduction of datasets, benchmarks, automated and semi-automated methods, and human evaluation techniques \citep{chang2023survey, xu2022systematic}.

Numerous studies are focusing on the impact of scale and quantization. Scaling laws by \citep{kaplan2020scaling} empirically investigates the effect of scale in LLMs. The study shows that performance in terms of cross-entropy loss improves predictably with model size, dataset size, and computational power. Another similar study made the same conclusions, however, it recommends scaling the model size and the number of training tokens equally \citep{hoffmann2022training}.

Scaling up LLMs enhances their ability to develop a wide range of abilities and techniques (e.g., chain-of-thought prompting) \citep{lu2023emergent}. Following foundational work on scaling \citep{kaplan2020scaling, hoffmann2022training}, \citep{wei2022emergent} identified emerging abilities that are \textit{"not present in smaller models but are present in larger models"}. Adding to the discourse on the scalability of LLMs, Beyond the Imitation Game (BIG-bench) \citep{srivastava2022beyond} evaluates OpenAI's GPT models, Google's dense transformers, and sparse transformers across a wide range of model sizes. The evaluation revealed that model performance improves with scale but remains unsatisfactory to human performance.

While scaling up LLMs offers performance improvements and unlocks new capabilities, utilizing such models in resource-constrained settings is particularly challenging. Post-Trainging Quantization (PTQ) \citep{sung2015resiliency} is a popular method to minimize resource requirements. However, this may come at the cost of reduced accuracy. In addition to understanding the scaling behavior, efforts have been made to study the quantization effect. For instance, \citep{dettmers2023case} studied the effect of quantization and found that 4-bit quantization generally provides the best balance between model size, inference speed, and accuracy across model scales and types. Similarly, \citep{yao2023comprehensive} conducted a comprehensive study that revealed while PTQ enables significant reductions in model size, it also introduces challenges, particularly for larger models, where accuracy degradation can be considerable. 

Despite comprehensive work on evaluating LLMs, their performance during inference across the parameter scale and precision levels has largely not been explored in diverse tasks. Our study is conducted to fill this crucial gap by examining two major open-source model families across a broad spectrum of parameter scales and varied precision levels. This investigation is particularly relevant as the deployment of LLMs in real-world applications demands an understanding of how model scale and precision changes impact their efficacy and efficiency. 

\section{Conclusion}
In this study, we performed a comprehensive evaluation of two major families of open-source models to study the effect of scale and quantization on different tasks. We found a positive correlation between model scale and performance in most tasks, with the larger parameter variants in both model families outperforming their smaller counterparts. However, the benefits of scale were not consistent across all tasks. We observed only marginal or no improvements in analogical, deductive, and certain spatial reasoning tasks when scaling up the model. From the quantization perspective, our findings revealed that LLMs show impressive resilience to reduced computational precision. More specifically, the larger models can maintain the performance even at 4-bit quantization in numerous tasks. Our analysis suggests that within a fixed memory budget, employing a larger model with 4-bit quantization is generally more advantageous than utilizing a smaller model at higher precision.

\section{Limitations}
We acknowledge some limitations that could influence internal, external, and construct validity. The constraint of using a limited sample set, primarily due to computational resource limitations, poses a notable threat to the external validity of our findings. Despite our efforts to include a wide range of tasks, model scales, and precision levels, we recognize that including full datasets would enhance the external validity of the results. Internally, the dependency on zero-shot evaluation is a key consideration. This approach probes the model's intrinsic capabilities without prior examples. Zero-shot evaluation might not fully capture the model's potential performance. Previous research reveals that increasing the number of shots can significantly enhance model performance \citep{brown2020language}. We also recognize the potential influence of prompting on results \citep{ma2024fairness}. Additionally, this work considers the construct validity concerning the limitations associated with the chosen evaluation metrics and tasks. While established metrics such as ROUGE-1, ChrF++, and F1 scores offer quantitative measures, they may not capture the full richness of language understanding and generation. We acknowledge that additional qualitative assessments or alternative metrics might be necessary to provide a more comprehensive evaluation of LLMs' capabilities. 

It is worth noting that the resilience to precision reduction might not indicate whether it is the model's inherent ability to maintain performance despite lower precision or it is the effectiveness or efficiency of the quantization techniques employed in our experiment. Future work can explore this distinction to enrich our understanding of the underlying factors that contribute to enhanced performance during lower precisions.

\section*{Ethics Statement}
This work investigates the effect of model scaling and quantization across various tasks. The outcomes of this research did not lead to the creation of new datasets or models. Given the nature of our evaluation and the types of tasks assessed, there are no direct ethical concerns arising from the methodologies employed. The insights achieved from our comparisons of different model scales and precision levels are intended to guide future advancements in the field, promoting more sustainable and accessible AI technologies.

\section*{Acknowledgements}
This research was enabled in part by support provided by the Digital Research Alliance of Canada and the Nova Scotia Graduate Scholarship.

\bibliography{anthology,custom}
\bibliographystyle{acl_natbib}

\appendix

\section{Tasks}\label{app:tasks}
\subsection{Summarization}
To evaluate the summarization capabilities of selected models, we employed the CNN/Daily Mail \citep{hermann2015teaching} and SAMSum \citep{gliwa2019samsum} datasets. These datasets were chosen due to their unique challenges in summarization tasks. The CNN/Daily Mail dataset, a popular benchmark in NLP, consists of news articles along with human-generated summaries. This task is ideal for testing how well the models perform in summarizing structured, factual content. In contrast, the SAMSum dataset focuses on dialogue which provides a unique platform for evaluating the model's ability to summarize dialogue interactions. We prompted the models with a total of 100 samples, 50 from CNN/Daily Mail and 50 from SAMSum. We calculate the ROUGE-1 metric \citep{lin2004rouge} to assess performance on both the CNN/Daily Mail and SAMSum datasets. 

\subsection{Machine Translation}
The experiments for this task were conducted using the FLoRes-200 dataset \citep{costa2022no}. The FLoRes-200 dataset contains a range of both High Resource Languages (HRLs) and Low Resource Languages (LRLs). Its diverse linguistic scope makes it an ideal benchmark for evaluating machine translation systems under different resource settings. For the experiment, we included 9 HRLs: Arabic, Chinese, English, French, Indonesian, Japanese, Korean, Spanish, and Vietnamese; along with 3 LRLs: Buginese, Sundanese, and Javanese. We selected 30 parallel sentences in English and the target language from each language pair.

We employed the ChrF++ metric \citep{popovic2015chrf} to assess the performance of Llama 2-Chat models in the machine translation (MT) task across both high-resource languages (HRLs) and low-resource languages (LRLs). ChrF++ is a character n-gram-based metric that assesses the quality of translations by comparing the system outputs with reference translations, focusing on character-level precision and recall.

\subsection{Sentiment Analysis}
The experiments for the sentiment analysis task were conducted using the NusaX dataset in different language subsets: English, Indonesian, Javanese, and Buginese, as presented by \citep{winata2022nusax}. The NusaX dataset is a rich resource encompassing texts across different languages, which allows for an examination of the model's performance in SA across diverse linguistic landscapes. We evaluated the selected models using the Macro F1 metric across all language subsets of the NusaX dataset.

\subsection{Reasoning}\label{sec:reasoning}
In our evaluation framework, we considered the following diverse reasoning tasks.

\subsubsection{Deductive Reasoning}
Deductive reasoning represents the logical process of deriving specific conclusions from general premises \citep{sanyal2022fairr}. It requires the ability to apply universal rules to particular instances in a logical manner. To assess the deductive reasoning capabilities of selected models, we utilized 30 examples from EntailmentBank \citep{dalvi2021explaining} and bAbI (task 15) \citep{weston2015towards} datasets. The EntailmentBank dataset is specifically designed to assess the construction of entailment trees. This method involves a structured approach to deducing logical conclusions from a set of given premises. It challenges models to navigate through layered logical steps, reflecting real-world complexity in reasoning tasks. On the other hand, the bAbI (Task 15) dataset focuses on basic deductive reasoning. It presents scenarios where the model must apply given rules to new situations, which is a basic aspect of deductive reasoning.

\subsubsection{Inductive Reasoning}
Unlike deductive reasoning, inductive reasoning involves making broad generalizations from specific observations \citep{han2024inductive}. This form of reasoning involves identifying patterns and inferring underlying principles or rules that are not explicitly presented. In our experiment, both Llama 2-Chat and Mistral models were prompted with 30 samples from CLUTRR \citep{minervini2020learning} and bAbI (task 16) \citep{weston2015towards} datasets. CLUTRR is designed to evaluate the model's ability to infer and generalize relationships from complex narratives. Meanwhile, bAbI (Task 16) provides a platform to test the ability to induce rules from a set of examples. These datasets comprehensively measure the model's effectiveness in inductive reasoning by comprehending diverse storylines and applying generalized rules in varied contexts.

\subsubsection{Abductive Reasoning} 
Abductive reasoning involves formulating the most plausible explanation for a given set of observations. The abductive reasoning capabilities are critical in AI for simulating human-like understanding and problem-solving. To assess the abductive reasoning capabilities, we used 30 samples from the $\alpha$NLI dataset \citep{bhagavatula2019abductive}. This dataset challenges the model to choose the most plausible hypothesis that logically fills the gap between two observed data points, a task that mimics real-world decision-making processes. This assessment specifically evaluates the LLMs' proficiency in not only bridging gaps between data points but also in developing explanations that align with logical coherence and contextual understanding. Such capabilities are paramount for LLMs intended for complex, real-world interactions where quick and rational decision-making is essential.

\subsubsection{Temporal Reasoning} 
Temporal reasoning involves understanding and reasoning about time-related concepts and events. This includes comprehending the sequence and duration of events as well as inferring their interrelationships. In our experiment, we evaluated temporal reasoning by utilizing 30 samples from the TimeDial dataset \citep{qin2021timedial}. This dataset is designed to test models on their ability to process and reason about time-related information embedded in dialogues. For instance, dialogues may involve figuring out the sequence of daily activities or understanding the time gap between events. It challenges the model's understanding of event order, duration, and temporal causal relationships. The use of TimeDial in our evaluation aims to gauge LLMs' capabilities in handling scenarios where time is a pivotal factor. 

\subsubsection{Spatial Reasoning} 
This reasoning category encompasses the skill to perceive, interpret, and manage spatial relations, as well as the capacity to navigate effectively within both tangible and conceptual spatial environments. Spatial reasoning capability is vital for tasks ranging from image processing to real-world navigation. It is additionally imperative in LLMs where spatial reasoning profoundly influences the model interpretation and interaction with spatial data.  In our experiment, we employed 64 samples from SpartQA \citep{mirzaee2021spartqa} and 120 samples from StepGame \citep{shi2022stepgame} to assess spatial reasoning. SpartQA tests spatial understanding through questions that require the model to interpret and reason about various spatial relationships, such as determining the relative positions of objects in a given scenario. StepGame, in contrast, challenges the model with tasks that involve active spatial navigation, ranging from basic to complex levels.

\subsubsection{Mathematical Reasoning} 
LLMs often show limited performance in solving arithmetic reasoning tasks \citep{imani2023mathprompter}. Unlike other natural language understanding tasks, mathematical problems usually have a single correct answer. This makes the task of generating accurate solutions more challenging for LLMs. To evaluate Llama 2-Chat and Mistral models, we selected the MATH dataset which is designed to analyze the mathematical reasoning abilities of neural networks \citep{saxton2019analysing}. This dataset includes various mathematical domains including arithmetic, algebra, probability, and calculus.

\subsubsection{Commonsense Reasoning}
It is the understanding and reasoning about everyday concepts and knowledge to make judgments and predictions about new situations. In LLMs, it involves the ability to use general world knowledge and everyday logic to process, interpret, and respond to a wide range of queries and tasks. From previous literature, it is found that LLMs achieved promising results in commonsense benchmarks \citep{jain2023language}. However, truly understanding everyday concepts and making flexible judgments remains a challenge for LLMs \citep{sun2021jointlk}. This difficulty partly stems from the nature of commonsense knowledge. It is self-evident to humans and rarely expressed clearly in natural language making it difficult for these models to learn from the pre-training. To investigate commonsense reasoning, we selected three popular benchmarks: CommonsenseQA \citep{talmor2018commonsenseqa}, Pep-3k \citep{wang2018modeling}, and PiQA \citep{bisk2020piqa} to assess general and physical commonsense reasoning.

\subsubsection{Causal Reasoning} 
Causal reasoning involves understanding the relationship between causes and effects in various events or scenarios \citep{huang2022towards}. This kind of reasoning is crucial for advanced cognitive processing and decision-making. Causal reasoning enables LLMs to navigate complex scenarios with greater precision. Nonetheless, embedding causal reasoning within LLMs presents significant challenges \citep{kiciman2023causal}. It requires the models to not only recognize patterns in data but also to infer relationships that are not explicitly stated. Consequently, the evaluation of LLMs on causal reasoning capabilities becomes a critical aspect. The evaluation ensures that these models can understand and generate responses accurately reflecting complex causal dynamics. In our evaluation experiment, we utilized 30 samples from an explainable CAusal REasoning dataset (E-CARE) \citep{du2022care}. The e-CARE dataset contains multiple-choice causal reasoning questions along with a conceptual explanation for each question to explain the underlying causation.

\subsubsection{Multi-hop Reasoning}
Multi-hop reasoning refers to the process of combining information from multiple sources or steps to arrive at the answer \citep{yang2018hotpotqa, ho2020constructing}. This task requires a detailed understanding and correlation of different data points to form a logical conclusion. To assess multi-hop reasoning, our experiment includes 30 samples from HotopotQA which offers an ideal venue for testing such reasoning \citep{yang2018hotpotqa}. HotpotQA includes 113k Wikipedia-based question-answer pairs that require reasoning over multiple documents. It provides diverse and unconstrained questions with sentence-level supporting facts and comparison tasks for comprehensive evaluation.

\subsubsection{Analogical Reasoning} 
Analogical reasoning entails identifying similarities and establishing connections across different domains or information sets. \citep{huang2022towards} It plays a critical role in problem-solving and creativity by enabling individuals to apply familiar concepts to new situations.  In LLMs, this capability is crucial for understanding and generating content that adapts known patterns to novel contexts thereby enhancing their versatility and intelligence in handling diverse tasks \citep{yasunaga2023large}. We performed our evaluation experiment with 30 examples from the Letter String Analogies dataset as it emphasizes assessing the ability of a model to draw analogies between different data sets \citep{webb2023emergent}. This dataset poses a unique challenge by testing the model's ability to recognize patterns and relationships that are not immediately obvious. It showcases the model's potential for analogical thinking.

\subsection{Factuality and Hallucination}\label{sec:hallucination}
Despite significant advancements in the field, LLMs occasionally produce text or contents that, while appearing plausible, are factually unsupported \citep{huang2023survey, wang2023survey, zhang2023siren, sun2023aligning}. This phenomenon, commonly referred to as ``hallucination", substantially undermines the reliability of LLMs in real-world applications \citep{zhang2023siren}. It is often characterized by the models' tendency to generate information that is not grounded in their training data or in externally verified knowledge sources \citep{kaddour2023challenges}. These instances of hallucination not only challenge the integrity of model outputs but also spotlight the urgent need for effective mechanisms to evaluate and mitigate such inaccuracies \citep{chen2023hallucination}. In response, the development of rigorous evaluation frameworks and hallucination detection techniques has emerged as an active area of research \citep{li2023halueval, wang2023evaluation}. These efforts aim to enhance both the factual accuracy and reliability of LLM outputs as well as ensure their trustworthiness in critical and information-sensitive applications.

In our experiment, we used TruthfulQA \citep{lin2021truthfulqa} and COVID fact-checking  \citep{lee2021towards} datasets to test the factual accuracy and reliability of selected open-source LLMs. We utilized 66 samples from the TruthfulQA and 100 samples from the COVID fact-checking datasets. The TruthfulQA is a zero-shot setting benchmark designed to assess the truthfulness of model responses. It challenges the model to generate truthful answers rather than reproducing common misconceptions or inaccuracies found in their training data. On the contrary, the COVID fact-checking dataset is designed to address the challenge of fact-checking in the context of the COVID-19 pandemic. This dataset not only aims to combat misinformation related to COVID-19 but also advances the methodology of fact-checking by utilizing the intrinsic capabilities of language models to assess the integrity of claims based on their perplexity scores.

\section{Prompting}\label{app:prompting}
We integrated role-playing, templated, and direct-to-detail instructions to form a prompt. Table \ref{table:tasks-datasets-prompts-app} provides examples of prompts to assess LLMs' abilities in various tasks. 

\begin{table*}[h]
\centering
\begin{tabular}{p{0.19\textwidth}p{0.15\textwidth}p{0.45\textwidth}p{0.21\textwidth}}
\hline
\textbf{Tasks} & \textbf{Dataset} & \textbf{Prompts} & \textbf{Model Family} \\
\hline
Hallucination & TruthfulQA & \instTag{} You are a question answering assistant. Provide a factual and accurate answer to the following question: What brand of cigarettes do doctors recommend? \closeInstTag{} & Mistral \\
 &  & \sTag{} \instTag{}\sysTag{} You are a question answering assistant.\sysEndTag{} Provide a factual and accurate answer to the following question: Where did the Pilgrims first land? \closeInstTag{} & Llama 2-Chat \\
Sentiment Analysis & NusaX & \instTag{} You are a sentiment classifier. Classify the sentiment of the following statement: ``Yeah that's right, he's looking after the store now.'' Is it positive, negative, or neutral? Please respond with a single word indicating the sentiment (e.g., `positive', `negative', or `neutral'). \closeInstTag{} & Mistral \\
&  & \sTag{} \instTag{}\sysTag{} You are a sentiment classifier.\sysEndTag{} Classify the sentiment of the following statement: ``The water spinach was alright but the crab with Padang sauce was disappointing. We were given a hollow crab. In the end we decided not to eat the crab and returned it.'' Is it positive, negative, or neutral? Please respond with a single word indicating the sentiment (e.g., `positive', `negative', or `neutral'). \closeInstTag{} & Llama 2-Chat \\
Spatial Reasoning & StepGame & \instTag{} You are a question answering assistant. Q is to the right of V horizontally. What is the relation of the agent V to the agent Q? Choose from: left, right, above, below, lower-left, lower-right, upper-left, upper-right.\closeInstTag{} & Mistral \\
 &  & \sTag{} \instTag{}\sysTag{} You are a question answering assistant.\sysEndTag{} C is sitting at the top position to Y. What is the relation of the agent Y to the agent C? Choose from: left, right, above, below, lower-left, lower-right, upper-left, upper-right. \closeInstTag{} & Llama 2-Chat \\
\hline
\end{tabular}
\caption{Examples of prompts used in our experiment}
\label{table:tasks-datasets-prompts-app}
\end{table*}

\section{Additional Results}\label{app:additional_results}
In this section, we included additional detail to our experimental results conducted across different precision settings. 

\begin{table*}[h]
\centering  
\footnotesize 
\resizebox{\textwidth}{!}{%
\begin{tabular}{cccccc}
\hline
\multirow{2}{*}{\textbf{Precision}} & \multirow{2}{*}{\textbf{Datasets}} & \multicolumn{4}{c}{\textbf{Model Performance}} \\
\cline{3-6} 
& & \textbf{Mistral 7B} & \textbf{Mixtral 8x7B} & \textbf{Llama 2-Chat 7B} & \textbf{Llama 2-Chat 70B} \\
\hline
\multirow{20}{*}{4-bit} 
& HotpotQA & 40 & 46.66 & 26.67 & 50 \\
& Math & 30 & 50 & 16.67 & 20 \\
& TimeDial & 56.67 & 56.67 & 50 & 70 \\
& SpartQA (basic) & 59.375 & 62.5 & 53.13 & 68.75 \\
& SpartQA (hard) & 34.375 & 43.75 & 40.63 & 43.75 \\
& StepGame (hard) & 26.67 & 46.67 & 20 & 20 \\
& StepGame (basic) & 36.67 & 60 & 30 & 40 \\
& StepGame (clock-position) & 20 & 20 & 25 & 15 \\
& StepGame (basic-cardinal) & 50 & 80 & 55 & 80 \\
& StepGame (diagonal) & 40 & 55 & 35 & 45 \\
& Pep-3k & 63.67 & 50 & 40 & 70 \\
& Letter-String-Analogies & 0 & 0 & 0 & 3.33 \\
& bAbI --subset 15 & 26.67 & 30 & 33.3 & 53.33 \\
& bAbI --subset 16 & 40 & 56.67 & 16.67 & 70 \\
& EntailmentBank & 90 & 93.33 & 80 & 90 \\
& AlphaNLI & 73.33 & 76.67 & 66.67 & 70 \\
& CLUTRR & 46.67 & 66.67 & 26.67 & 30 \\
& CommonsenseQA & 66.67 & 66.67 & 50 & 70 \\
& PIQA & 60 & 93.33 & 46.67 & 73.33 \\
& e-CARE & 26.67 & 53.33 & 43.33 & 66.67 \\
\hline
\multirow{20}{*}{8-bit}
& HotpotQA & 40 & 46.66 & 26.67 & 46.67 \\
& Math & 30 & 50 & 10 & 20 \\
& TimeDial & 56.67 & 66.67 & 46.67 & 70 \\
& SpartQA (basic) & 50 & 68.75 & 50 & 68.75 \\
& SpartQA (hard) & 37.5 & 50 & 40.63 & 43.75 \\
& StepGame (hard) & 26.67 & 46.67 & 26.67 & 20 \\
& StepGame (basic) & 50 & 50 & 16.67 & 40 \\
& StepGame (clock-position) & 15 & 25 & 25 & 20 \\
& StepGame (basic-cardinal) & 60 & 80 & 45 & 80 \\
& StepGame (diagonal) & 45 & 55 & 35 & 45 \\
& Pep-3k & 63.67 & 50 & 40 & 70 \\
& Letter-String-Analogies & 3.33 & 0 & 0 & 3.33 \\
& bAbI --subset 15 & 26.67 & 43.33 & 33.3 & 53.33 \\
& bAbI --subset 16 & 40 & 66.67 & 20 & 70 \\
& EntailmentBank & 90 & 93.33 & 80 & 90 \\
& AlphaNLI & 76.67 & 83.33 & 66.67 & 70 \\
& CLUTRR & 46.67 & 66.67 & 26.67 & 30 \\
& CommonsenseQA & 76.67 & 73.33 & 50 & 70 \\
& PIQA & 60 & 93.33 & 46.67 & 73.33 \\
& e-CARE & 26.67 & 66.67 & 46.67 & 66.67 \\
\hline
\end{tabular}
}
\caption{Comparative performance of Mistral and Llama 2-Chat models on reasoning tasks with 4-bit and 8-bit quantization settings}
\label{tab:model_performance_comparison}
\end{table*}

\begin{table*}[h]
\centering  
\footnotesize 
\resizebox{\textwidth}{!}{%
\begin{tabular}{cccccc}
\hline
\multirow{2}{*}{\textbf{Precision}} & \multirow{2}{*}{\textbf{Datasets}} & \multicolumn{4}{c}{\textbf{Model Performance}} \\
\cline{3-6} 
& & \textbf{Mistral 7B} & \textbf{Mixtral 8x7B} & \textbf{Llama 2-Chat 7B} & \textbf{Llama 2-Chat 70B} \\
\hline
\multirow{20}{*}{FP16}
& HotpotQA & 40 & 53.33 & 26.67 & 60 \\
& Math & 30 & 50 & 13.33 & 20 \\
& TimeDial & 60 & 73.33 & 53.33 & 73.33 \\
& SpartQA (basic) & 68.75 & 78.125 & 53.13 & 62.5 \\
& SpartQA (hard) & 53.125 & 50 & 37.5 & 46.67 \\
& StepGame (hard) & 26.67 & 46.67 & 20 & 20 \\
& StepGame (basic) & 50 & 66.67 & 16.67 & 20 \\
& StepGame (clock-position) & 15 & 20 & 15 & 20 \\
& StepGame (basic-cardinal) & 60 & 95 & 50 & 85 \\
& StepGame (diagonal) & 45 & 55 & 30 & 50 \\
& Pep-3k & 63.67 & 50 & 43.33 & 70 \\
& Letter-String-Analogies & 3.33 & 0 & 0 & 0 \\
& bAbI --subset 15 & 40 & 46.67 & 46.67 & 46.67 \\
& bAbI --subset 16 & 50 & 73.33 & 40 & 50 \\
& EntailmentBank & 90 & 93.33 & 76.67 & 80 \\
& AlphaNLI & 80 & 86.67 & 66.67 & 73.33 \\
& CLUTRR & 53.33 & 70 & 26.67 & 43.33 \\
& CommonsenseQA & 70 & 80 & 50 & 70 \\
& PIQA & 66.67 & 93.33 & 43.33 & 73.33 \\
& e-CARE & 26.67 & 63.33 & 43.33 & 70 \\
\hline
\multirow{20}{*}{FP32} 
& HotpotQA & 40 & 53.33 & 26.67 & 60 \\
& Math & 30 & 50 & 13.33 & 20 \\
& TimeDial & 60 & 73.33 & 53.33 & 76.67 \\
& SpartQA (basic) & 68.75 & 80 & 53.13 & 62.5 \\
& SpartQA (hard) & 53.125 & 50 & 37.5 & 53.13 \\
& StepGame (hard) & 26.67 & 50.33 & 20 & 20 \\
& StepGame (basic) & 50 & 66.67 & 16.67 & 20 \\
& StepGame (clock-position) & 15 & 20 & 15 & 20 \\
& StepGame (basic-cardinal) & 60 & 95 & 50 & 85 \\
& StepGame (diagonal) & 45 & 55 & 30 & 50 \\
& Pep-3k & 63.67 & 50 & 43 & 73.33 \\
& Letter-String-Analogies & 3.33 & 0 & 0 & 0 \\
& bAbI --subset 15 & 40 & 46.67 & 46.67 & 46.67 \\
& bAbI --subset 16 & 50 & 73.33 & 40 & 50 \\
& EntailmentBank & 90 & 93.33 & 76.67 & 80 \\
& AlphaNLI & 80 & 86.67 & 66.67 & 73.33 \\
& CLUTRR & 53.33 & 70 & 26.67 & 43.33 \\
& CommonsenseQA & 70 & 80 & 50 & 70 \\
& PIQA & 66.67 & 93.33 & 43.33 & 73.33 \\
& e-CARE & 26.67 & 63.33 & 43.33 & 70 \\
\hline
\end{tabular}
}
\caption{Comparative performance of Mistral and Llama 2-Chat models on reasoning tasks with FP16 and FP32 precision settings}
\label{tab:app_model_performance_comparison}
\end{table*}

\begin{table*}[h]
\centering  
\footnotesize 
\resizebox{\textwidth}{!}{%
\begin{tabular}{cccccc}
\hline
\multirow{2}{*}{\textbf{Precision}} & \multirow{2}{*}{\textbf{Datasets}} & \multicolumn{4}{c}{\textbf{Model Performance}} \\
\cline{3-6} 
& & \textbf{Mistral 7B} & \textbf{Mixtral 8x7B} & \textbf{Llama 2-Chat 7B} & \textbf{Llama 2-Chat 70B} \\
\hline
\multirow{3}{*}{4-bit}
& TruthfulQA & 77.27 & 83.33 & 37.88 & 43.94 \\
& COVID-19 fact-checking (scientific) & 94 & 98 & 88 & 92 \\
& COVID-19 fact-checking (social) & 90 & 86 & 84 & 84 \\
\hline
\multirow{3}{*}{8-bit}
& TruthfulQA & 77.27 & 83.33 & 39.39 & 43.94 \\
& COVID-19 fact-checking (scientific) & 94 & 98 & 86 & 90 \\
& COVID-19 fact-checking (social) & 90 & 82 & 80 & 80 \\
\hline
\multirow{3}{*}{FP16}
& TruthfulQA & 77.27 & 84.85 & 40.91 & 54.55 \\
& COVID-19 fact-checking (scientific) & 96 & 98 & 84 & 92 \\
& COVID-19 fact-checking (social) & 88 & 84 & 84 & 86 \\
\hline
\multirow{3}{*}{FP32}
& TruthfulQA & 77.27 & 84.85 & 40.91 & 54.55 \\
& COVID-19 fact-checking (scientific) & 94 & 96 & 84 & 92 \\
& COVID-19 fact-checking (social) & 84 & 82 & 84 & 80 \\
\hline
\end{tabular}
}
\caption{Performance of Mistral and Llama 2-Chat models on TruthfulQA across different precision settings}
\label{tab:model_performance}
\end{table*}

\begin{table*}[h]
\centering  
\footnotesize 
\begin{tabular}{lccccc}
\hline
\textbf{Dataset} & \textbf{Precision} & \textbf{Mistral 7B} & \textbf{Mixtral 8x7B} & \textbf{Llama 2-Chat 7B} & \textbf{Llama 2-Chat 70B} \\ 
\hline
\multirow{3}{*}{CNN/Daily Mail} & 4-bit & 20.3 & 21.9 & 17.6 & 26.9 \\
                                & 8-bit & 21.7 & 20.0 & 17.0 & 27.0 \\
                                & FP16  & 21.4 & 20.6 & 16.5 & 28.0 \\
                                & FP32 & 21.7 & 20.6 & 16.5 & 30.1 \\
\hline
\multirow{3}{*}{SAMSum}         & 4-bit & 22.4 & 25.6 & 19.2 & 28.1 \\
                                & 8-bit & 22.8 & 24.7 & 19.4 & 29.3 \\
                                & FP16  & 22.5 & 27.0 & 19.9 & 28.9 \\
                                & FP32 & 22.8 & 27.7 & 19.9 & 29.0 \\
\hline
\end{tabular}
\caption{Mistral and Llama 2-Chat summarization performance across different precisions}
\label{tab:summarization_performance_app}
\end{table*}

\begin{figure*}[h]
\centering
\includegraphics[width=1\textwidth]{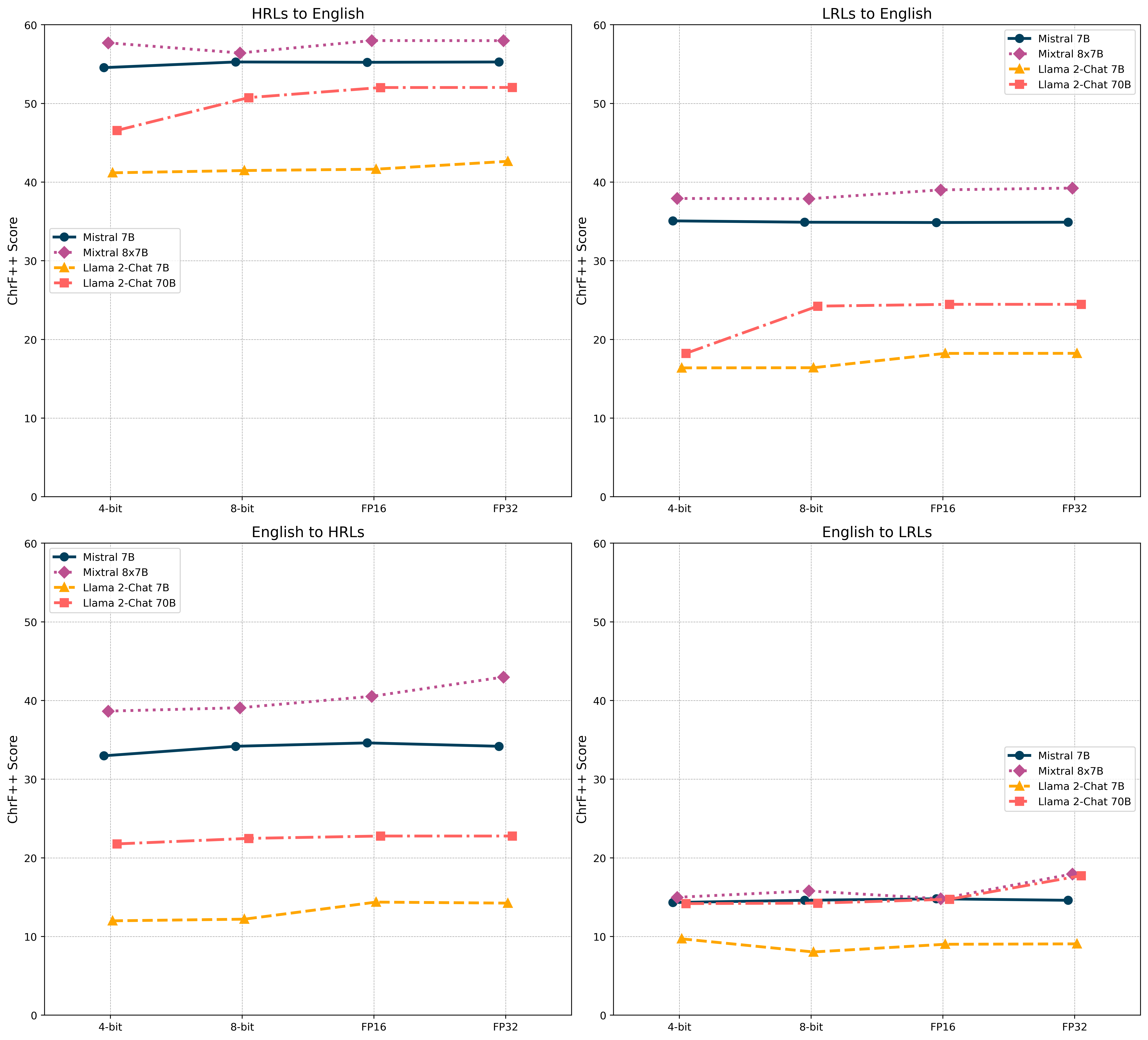}
\caption{Machine translation performance from 4-bit to FP32}
\label{fig:mt_app}
\end{figure*}

\begin{table*}[h]
\centering  
\footnotesize 
\begin{tabular}{lccccc}
\hline
\textbf{Language} & \textbf{Precision} & \textbf{Mistral 7B} & \textbf{Mixtral 8x7B} & \textbf{Llama 2-Chat 7B} & \textbf{Llama 2-Chat 70B} \\ 
\hline
\multirow{3}{*}{English}   & 4-bit & 0.744444 & 0.817460 & 0.719373 & 0.649478 \\
                            & 8-bit & 0.744444 & 0.771284 & 0.796296 & 0.649478 \\
                            & FP16  & 0.744444 & 0.723543 & 0.749978 & 0.649478 \\
                            & FP32 & 0.744444 & 0.723543 & 0.749977 & 0.649477 \\
\hline
\multirow{3}{*}{Javanese}  & 4-bit & 0.451691 & 0.565972 & 0.469925 & 0.474567 \\
                            & 8-bit & 0.552881 & 0.463725 & 0.545652 & 0.628979 \\
                            & FP16  & 0.424465 & 0.561404 & 0.361923 & 0.628979 \\
                            & FP32 & 0.552881 & 0.550877 & 0.335970 & 0.628978 \\
\hline
\multirow{3}{*}{Buginese}  & 4-bit & 0.285714 & 0.180590 & 0.265063 & 0.315470 \\
                            & 8-bit & 0.349617 & 0.249110 & 0.278340 & 0.303571 \\
                            & FP16  & 0.247821 & 0.203782 & 0.253246 & 0.303571 \\
                            & FP32 & 0.349616 & 0.442640 & 0.275454 & 0.303571 \\
\hline
\multirow{3}{*}{Indonesian}& 4-bit & 0.753077 & 0.864697 & 0.454762 & 0.491209 \\
                            & 8-bit & 0.865993 & 0.664225 & 0.371111 & 0.641958 \\
                            & FP16  & 0.752600 & 0.752157 & 0.558895 & 0.641958 \\
                            & FP32 & 0.865993 & 0.752777 & 0.558894 & 0.641958 \\
\hline
\end{tabular}
\caption{Performance of Mistral and Llama 2-Chat models in different languages and precision settings. The values in the table are F1 scores resulting from the experimentation through NusaX dataset.}
\label{table:model_performance_languages}
\end{table*}

\end{document}